\documentclass{article}

\PassOptionsToPackage{square,sort,comma,numbers}{natbib}

\usepackage[preprint]{neurips_2024}

\usepackage[utf8]{inputenc} 
\usepackage[T1]{fontenc}    
\usepackage{hyperref}       
\usepackage{url}            
\usepackage{graphicx}
\usepackage{booktabs}       
\usepackage{multicol}
\usepackage{multirow}
\usepackage{amsfonts}       
\usepackage{nicefrac}       
\usepackage{microtype}      
\usepackage{xcolor}         
\usepackage{amsmath}
\usepackage[font=footnotesize]{caption}
\usepackage{subcaption}
\usepackage{scalerel}
\usepackage{svg}
\usepackage[utf8]{inputenc} 
\usepackage[T1]{fontenc}    
\usepackage{hyperref}       
\usepackage{url}            
\usepackage{booktabs}       
\usepackage{amsfonts}       
\usepackage{nicefrac}       
\usepackage{microtype}      
\usepackage{xcolor}         
\usepackage{graphicx}
\usepackage{amsmath}
\usepackage{ulem}
\usepackage{multirow}

\title{FPN-fusion: Enhanced Linear Complexity Time Series Forecasting Model}

\author{%
	Chu Li \\
	University of Science and Technology of China\\
	Hefei, China\\
	\texttt{lichuzm@mail.ustc.edu.cn} \\
	\And
	Bingjia Xiao \\
	Institute of Plasma Physics \\
	Hefei Institutes of Physical Science, Chinese Academy of Sciences\\
	Hefei, China \\
	\texttt{bjxiao@ipp.ac.cn} \\
	\and
	Qiping Yuan \\
	Institute of Plasma Physics \\
	Hefei Institutes of Physical Science, Chinese Academy of Sciences\\
	Hefei, China \\
	\texttt{qpyuan@ipp.ac.cn} \\
}

\begin{document}

	\maketitle

	\begin{abstract}
		This study presents a novel time series prediction model, FPN-fusion, designed with linear computational complexity, demonstrating superior predictive performance compared to DLiner without increasing parameter count or computational demands. Our model introduces two key innovations: first, a Feature Pyramid Network (FPN) is employed to effectively capture time series data characteristics, bypassing the traditional decomposition into trend and seasonal components. Second, a multi-level fusion structure is developed to integrate deep and shallow features seamlessly. Empirically, FPN-fusion outperforms DLiner in 31 out of 32 test cases on eight open-source datasets, with an average reduction of 16.8\% in mean squared error (MSE) and 11.8\% in mean absolute error (MAE). Additionally, compared to the transformer-based PatchTST, FPN-fusion achieves 10 best MSE and 15 best MAE results, using only 8\% of PatchTST's total computational load in the 32 test projects.
	\end{abstract}
	\section{Introduction}
	Time series forecasting, a critical area in deep learning research, finds extensive applications in finance \cite{mine_finance}, weather forecasting \cite{ltt_weather}, and sensor data analysis \cite{ztl3_sensor}, among others. The need for accurate long-term predictions in various domains has driven the development of sophisticated algorithms, including deep learning models like ARIMA \cite{arima}, GBRT, and recursive neural networks \cite{RNN1}, as well as more recent architectures like Causal Time Networks (TCNs) \cite{tcn} and the Mixer \cite{mixer}.
	
	Transformers, initially designed for NLP tasks, have also made significant strides in time series forecasting, with models like PatchTST \cite{PatchTST}, ETSformer \cite{ETSformer:}, Autoformer \cite{autoformer}, and FEDformer \cite{fedformer} showcasing their prowess. However, these models often come with high computational complexity and a large number of parameters. In contrast, DLlinear \cite{dlinear} offers a linear complexity alternative, decomposing data into trend and seasonal components and predicting them separately, achieving strong performance despite its simplicity.
	
	\begin{figure*}[ht]
		\begin{center}
			\centerline{\includegraphics[scale=0.5]{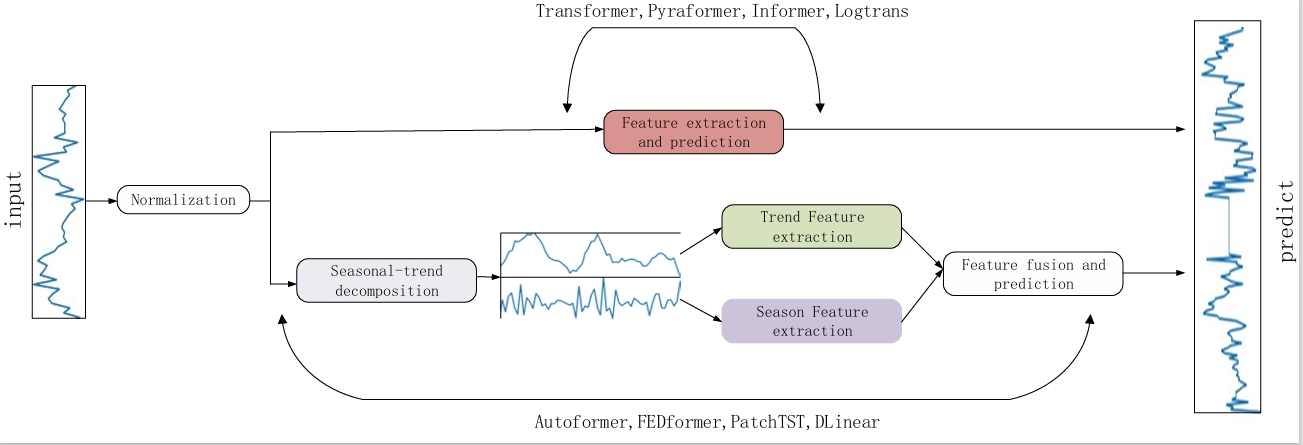}}
			\caption{The pipeline of existing Model TSF solutions.}
			\label{data-decompose}
		\end{center}
	\end{figure*}
	
	Data decomposition and normalization are the most commonly used methods to improve the predictive ability of models. The earliest origin of data decomposition is the ARMA\cite{arima}, which decomposes time series data into four parts: trend factor (T), cycle factor (C), seasonal factor (S), and random factor (I), which is more in line with human perception of data. Decomposing the data into trend and seasonal terms is a simplification of these four parts. As shown in Figure \ref{data-decompose},Traditional machine learning is based on raw data, extracting features from the raw data and predicting future data, such as GBRT, Informar\cite{informer}, DLinear\cite{dlinear},Logtrans\cite{logtrans},DeepAR\cite{deepar},LSTMnetwork\cite{LSTMnetwork},etc. There is another data decomposition strategy: to decompose time series data into trend and seasonal items. The PatchTST\cite{PatchTST} and DLlinear\cite{dlinear} decompose time series data into trend and seasonal terms, extract and predict the two separately, and add them up to output the prediction results. Autoformer\cite{autoformer} and FEDformer\cite{fedformer} use feature fusion models to fuse trend and seasonal features, and use the fused features to predict the future. After verification by many scientific researchers, binary decomposition is a very suitable method for deep learning. However, the data binary decomposition method has two problems: 
	\begin{itemize}
		\item The inherent relationship between trend and seasonal characteristics is overlooked in a crude decomposition. This separation can lead to the loss of trend information within seasonal components and vice versa, as the distinct characteristics are intertwined.
		\item Current approaches, which involve feature extraction for trend and seasonal terms separately, followed by simple addition for fusion, do not effectively account for the correlation between these features. This limits the model's ability to capture and utilize the complex interactions between trend and seasonality in the data. 
	\end{itemize}

	To address these limitations, this paper contributes a novel FPN-based multi-level fusion model, inspired by the U-Net architecture, for time series forecasting. Our contributions are two-fold:

	\begin{itemize}
		\item \textbf{Feature Pyramid Network (FPN) for time series analysis:} We leverage FPN \cite{FPN} to describe the time series data by applying multi-layer operations through pooling functions. This method replaces the conventional raw data approach and binary decomposition, offering a more efficient way to extract both shallow (seasonal and trend) and deep features. The use of small pooling kernels in FPN reduces computational complexity while enhancing the extraction of temporal features.
		\item  \textbf{FPN-fusion model:} Our proposed model integrates an FPN for feature extraction with a multi-level fusion module. FPN captures deep trend information, while low-level features capture seasonal elements. The fusion mechanism gradually combines these features, resulting in improved performance. Empirically, FPN-fusion outperforms DLiner \cite{dlinear} with 31 optimal mean absolute error (MAE) and mean squared error (MSE) scores, respectively, achieving an average decrease of 16.8\% in MSE and 11.8\% in MAE across 32 multivariate forecasting tasks. Against PatchTST \cite{PatchTST}, FPN-fusion exhibits 9 optimal MSE and 15 optimal MAE results.
	\end{itemize}
	
\begin{figure*}[ht]
		\centering
		\includegraphics[width=\textwidth]{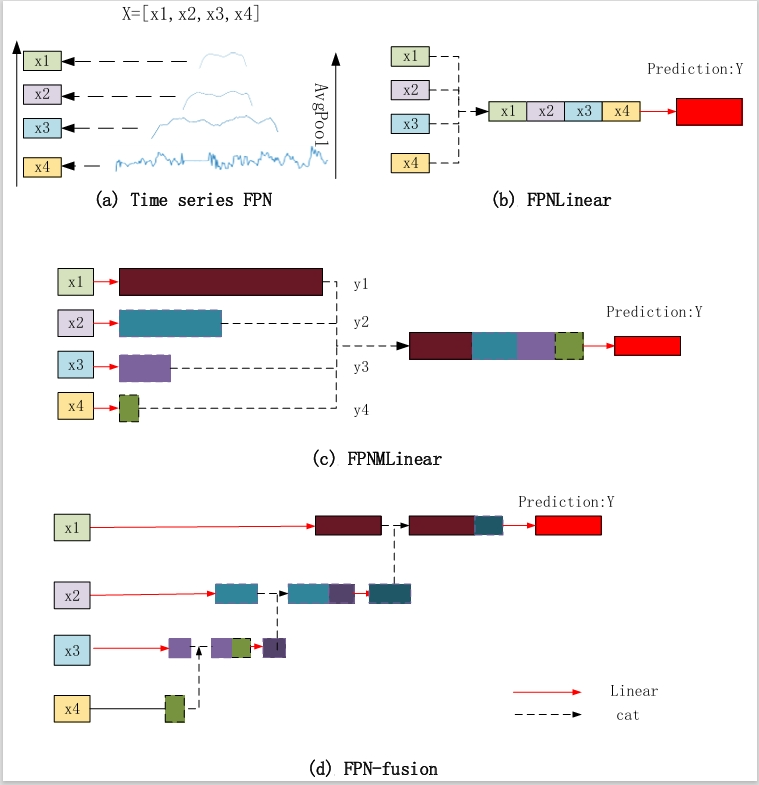}
		\caption{(a) Feature Pyramid Network for time series analysis. (b) FPNLinear algorithm structure. (c) FPNMLinear algorithm structure. (d) FPN-fusion algorithm structure.}
		\label{FPN-fusion architecture}
	\end{figure*}

	In Section Proposed Method, we present our methodology in detail and validate its effectiveness through extensive experiments. To analyze the FPN fusion model's components, we compare it with FPNLinear, FPNMLinear, NLinear, and DLiner models, using 8 publicly available datasets and 32 tests. Figure \ref{FPN-fusion architecture} illustrates the structural diagrams of the FPN feature pyramid, FPNLinear, FPNMLinear, and FPN fusion models.
	
	\section{Related work}
	\subsection{Linear Time Series Models}
	Linear time series models, characterized by a complexity of $O(L)$, encompass models such as ARMA\cite{arima}, NLinear, and DLlinear\cite{dlinear}. DLlinear\cite{dlinear} and NLlinear\cite{dlinear} are currently fundamental in long-term forecasting tasks. These models linearly normalize input data and employ fully connected networks. NLlinear employs the nearest value, $x_L$, as a base, and constructs a new historical data point, $x^{'}$, by subtracting $x - x_L$. This is fed into a fully connected layer, generating a prediction, $y^{'}$, which is then added to $x_L$ to yield the forecasted data, $y$. DLlinear\cite{dlinear} decomposes time series into seasonal-term, $x_{season}$, and trend-term, $x_{trend}$, using data decomposition techniques, and predicts each component separately with fully connected layers, summing the results for the final prediction. The overall complexity of these linear models remains at $O(L)$.
	
	\subsection{Transformer-Based Models}
	The Transformer's\cite{bert} advancements have inspired its application in long-term time series forecasting. Notable examples include LogTrans\cite{logtrans}, which utilizes convolutional self-attention layers and LogSparse design to capture local information and reduce spatial complexity. Informer\cite{informer} introduces ProbSparse self-attention for efficient feature extraction. Autoformer\cite{autoformer} combines time series analysis with Transformers, decomposing data into trend and seasonal components, and introduces autocorrelation. FEDformer\cite{fedformer} employs Fourier-enhanced structures for linear complexity, while Yformer\cite{Yformer} employs a U-Net-inspired Transformer design with downscaling and upsampling for long-range effects. Pyraformer\cite{Pyraformer} introduces a pyramid attention module with inter- and intra-scale connections, maintaining linear complexity. PatchTST\cite{PatchTST} applies patch techniques to shorten sequences, reducing complexity and enhancing local features.
	
	\subsection{Time Series Decomposition Techniques}
	A time series is composed of trends, seasonal fluctuations, cyclic variations, and irregular components. Long-term trends represent persistent changes over extended periods, seasonal fluctuations are regular patterns due to seasonal variations, cyclic fluctuations are periodic fluctuations without strict rules, and irregular fluctuations result from random influences. Autoformer\cite{autoformer} initiates the use of seasonal trend decomposition to enhance predictability by inputting trend components derived from the sequence. The residual between the original sequence and trend components is considered seasonal. FEDformer\cite{fedformer} extends this by proposing an expert strategy for mixed trend extraction, using moving average kernels with varying sizes. PatchTST\cite{PatchTST} and DLlinear\cite{dlinear} adopt binary seasonal trend decomposition, isolating seasonal and trend components for separate prediction, then combining them for the final forecast.
	
	\section{Proposed Method}
	The task of time series forecasting involves predicting {T} future data points based on a historical look-back window of fixed length L. Given input data $x = \{x_{1}^c, x_{2}^c, x_{3}^c, ..., x_{L}^c\}$, where each $x_{i}^c$ represents a vector dimension of channel $c$ at time $t = i$, and $c$ denotes the number of channels in the dataset, the model aims to output the forecasted data $y = \{x_{L+1}^c, x_{L+2}^c, ..., x_{L+T}^c\}$.
	
	\subsection{FPN of time series}
	In the realm of image analysis, the Feature Pyramid Network (FPN) \cite{FPN} effectively addresses the challenge of detecting objects at large scales by capturing features at diverse scales. Similarly, in the context of time series analysis, decomposing data into trend and seasonal components is a prevalent strategy for enhancing prediction accuracy. Inspired by FPN's success, we employ pooling operations to extract time series features, creating a feature pyramid. Deep layers contain more trend information, while shallow layers hold more seasonal details.
	
	For instance, when setting the data into four layers (with \( stage=4 \)), we utilize average pooling with a kernel size of 3, stride of 2, and padding of 0. Initially, we define the input data as \( x \). After passing through the FPN module, we obtain new input data \( X = [x_1, x_2, x_3, x_4] \), where:
	
	\[ x_1 = x \]
	\[ x_2 = AvgPool(x_1, kernel\_size=3, stride=2, padding=0) \]
	\[ x_3 = AvgPool(x_2, kernel\_size=3, stride=2, padding=0) \]
	\[ x_4 = AvgPool(x_3, kernel\_size=3, stride=2, padding=0) \]
	The length of each \( x_i \) is calculated as:
	\[{\rm{len(}}{{\rm{x}}_i}{\rm{)  = }}\left\lfloor {\frac{{x_{i - 1} + 2 \times padding - kernel\_size}}{{stride}} + 1} \right\rfloor \]
	
	As depicted in Figure \ref{FPN-fusion architecture}, the FPN architecture for time series effectively isolates trend features. The trend information in the top layer is more concentrated, while the seasonal aspects are more abundant in the lower layers.
	
	\subsection{FPNLinear and FPNMLinear}
	FPNLinear and FPNMLinear are introduced. FPNLinear concatenates FPN feature layers and employs a fully connected output for length prediction. FPNMLinear, on the other hand, applies fully connected layers to each FPN feature layer, generating scale-specific outputs \( Y = [y_1, y_2, y_3, y_4] \), where the length of each \( y_i \) mirrors that of \( x \) extraction using average pooling.
	
	\subsection{FPN-fusion model}
	The multi-stage fusion module takes advantage of the multi-scale time series feature \( X \) by concatenating \( y_i \) and \( y_{i-1} \), followed by a fully connected layer to produce \( y_i^{'} \), maintaining the same length as \( y_i \).
	
	The proposed FPN-fusion model, illustrated in Figure \ref{FPN-fusion architecture}, consists of two main components: time series FPN and a feature fusion module. It mirrors the structure of the U-Net network \cite{Unet}, combining fully connected and pooling layers. The left side processes time series FPN, forming a descriptive feature set \( X=\{X_1, X_2, X_3,...,X_{S}\} \) with \( S \) layers. The right side houses the fusion module, which integrates upper-layer features with local ones through fully connected layers, preserving feature length.
	
	\subsection{Evaluation metric}
	Performance evaluation is based on the widely-used Mean Squared Error (MSE) and Mean Absolute Error (MAE) \cite{mae}, as per previous research.
	
	\section{Experiments}

	In this study, we assess the efficacy of our proposed FPN-fusion model on eight real-world datasets, including ETT (ETTh1, ETTh2, ETTm1, ETTm2), transportation, electricity, weather, and ILI. These datasets, which are commonly employed for benchmarking and publicly available \cite{autoformer}, are chosen for their extensive use. Notably, we focus on larger datasets such as weather, transportation, ILI, and electricity, as they boast a larger number of time series, which aids in obtaining more stable and less prone to overfitting results. For univariate forecasting, ETT is employed, while multivariate forecasting is conducted on the remaining eight datasets. The statistical details of these datasets are summarized in Table \ref{tab:dataset}.
	
	\begin{table*}[htbp]
		\caption{The statistics of the nine popular datasets for benchmark.}
		\label{tab:dataset}
		\vskip 0.15in
		\begin{center}
			\begin{small}
				\begin{tabular}{l|ccccccccr}
					\toprule
					Datasets & ETTh1 & ETTh2 & ETTm1 & ETTm2 & traffic & Electricity & Weather & ILI  \\
					\midrule
					Variates   & 7 & 7 & 7 & 7& 862 &321 &211 &7  \\
					Timessteps & 17420 & 17420 & 69680 & 69680 & 17544 & 26304  &52696 &966\\
					Granularity & 1hour & 1hour & 5min & 5min & 1hour & 1hour  &10min &11week\\
					\bottomrule
				\end{tabular}
			\end{small}
		\end{center}
		\vskip -0.1in
	\end{table*}

	\subsection{Platform}
	the platform employed for experimentation was a Windows system, equipped with an Intel Xeon Gold 6226R CPU running at 2.90 GHz (dual-core) and a single Nvidia RTX A4000 graphics processing unit, offering 16 GB of memory. The section below presents the results and analysis of univariate long-term forecasting, using the ETT datasets with prediction lengths ranging from 96 to 720 time steps. The best performance for each metric is highlighted in bold for clarity.

	\subsection{Compared SOTA methods}
	To compare our model with state-of-the-art (SOTA) methods, we consider Transformer-based models like PatchTST \cite{PatchTST}, FEDformer \cite{fedformer}, Autoformer \cite{autoformer}, and Informer \cite{informer}. Additionally, we include the linear model DLiner \cite{dlinear} as a baseline. All models are tested under consistent experimental conditions, with prediction lengths \( T \) varying from 24 to 60 for the ILI dataset and from 96 to 720 for the others, and look-back windows \( L \) following the original paper specifications. The baseline results for PatchTST \cite{PatchTST} and DLiner \cite{dlinear} are obtained using the optimal configurations provided in their respective papers, with the PatchTST results taken from the best-performing configurations (patchTST/64 and patchTST/42).
	
	\begin{table*}[htbp]
		\caption{Univariate long-term forecasting results. ETT datasets are used with prediction lengths $T\in \{96, 192, 336, 720\}$. The best results are in \textbf{bold}.}
		\centering
		\label{tab::univariate}
		\vskip 0.15in
		\centering
		\resizebox{\linewidth}{!}{
			\begin{tabular}{cc|c|cc|cc|cc|cc|cc|cc|ccc}
				\cline{2-17}
				&\multicolumn{2}{c|}{Models}& \multicolumn{2}{c|}{FPN-fusion}& \multicolumn{2}{c|}{PatchTST}& \multicolumn{2}{c|}{DLinear}& \multicolumn{2}{c|}{FEDformer}& \multicolumn{2}{c|}{Autoformer}& \multicolumn{2}{c|}{Informer}& \multicolumn{2}{c}{LogTrans} \\
				\cline{2-17}
				&\multicolumn{2}{c|}{Metric}&MSE&MAE&MSE&MAE&MSE&MAE&MSE&MAE&MSE&MAE&MSE&MAE&MSE&MAE\\
				\cline{2-17}
				&\multirow{4}*{\rotatebox{90}{ETTh1}}& 96    & \textbf{0.055} & \textbf{0.178} & \textbf{0.055} & {0.179} & 0.056 & 0.180 & 0.079 & 0.215 & 0.071 & 0.206 & 0.193 & 0.377 & 0.283 & 0.468 \\
				&\multicolumn{1}{c|}{}& 192   & \textbf{0.071} & 0.205 & \textbf{0.071} & 0.205 & \textbf{0.071} & \textbf{0.204} & 0.104 & 0.245 & 0.114 & 0.262 & 0.217 & 0.395 & 0.234 & 0.409 \\
				&\multicolumn{1}{c|}{}& 336   & {0.078} & {0.221} & \textbf{0.076} & \textbf{0.220} & 0.098 & 0.244 & 0.119 & 0.270 & 0.107 & 0.258 & 0.202 & 0.381 & 0.386 & 0.546 \\
				&\multicolumn{1}{c|}{}& 720   & \textbf{0.087} & 0.233 & \textbf{0.087} & \textbf{0.232} & 0.189 & 0.359 & 0.142 & 0.299 & 0.126 & 0.283 & 0.183 & 0.355 & 0.475 & 0.629 \\
				\cline{2-17}
				&\multirow{4}*{\rotatebox{90}{ETTh2}}& 96    & \textbf{0.124} & \textbf{0.271} & {0.129} & 0.282 & 0.131 & {0.279} & 0.128 & 0.271 & 0.153 & 0.306 & 0.213 & 0.373 & 0.217 & 0.379 \\
				&\multicolumn{1}{c|}{}& 192   & \textbf{0.165} & \textbf{0.320} & {0.168} & {0.328} & 0.176 & 0.329 & 0.185 & 0.330 & 0.204 & 0.351 & 0.227 & 0.387 & 0.281 & 0.429 \\
				&\multicolumn{1}{c|}{}& 336   & {0.184} & {0.348} & \textbf{0.171} & \textbf{0.336} & 0.209 & 0.367 & 0.231 & 0.378 & 0.246 & 0.389 & 0.242 & 0.401 & 0.293 & 0.437 \\
				&\multicolumn{1}{c|}{}& 720   & {0.227} & {0.382} & \textbf{0.223} & \textbf{0.380} & 0.276 & 0.426 & 0.278 & 0.420 & 0.268 & 0.409 & 0.291 & 0.439 & 0.218 & 0.387 \\
				\cline{2-17}
				&\multirow{4}*{\rotatebox{90}{ETTm1}}& 96    & \textbf{0.026} & \textbf{0.121} & \textbf{0.026} & \textbf{0.121} & 0.028 & 0.123 & 0.033 & 0.140 & 0.056 & 0.183 & 0.109 & 0.277 & 0.049 & 0.171 \\
				&\multicolumn{1}{c|}{}& 192   & \textbf{0.039} & \textbf{0.149} & \textbf{0.039} & {0.150} & 0.045 & 0.156 & 0.058 & 0.186 & 0.081 & 0.216 & 0.151 & 0.310 & 0.157 & 0.317 \\
				&\multicolumn{1}{c|}{}& 336   & \textbf{0.051} & \textbf{0.171} & \textbf{0.053} & {0.173} & 0.061 & 0.182 & 0.084 & 0.231 & 0.076 & 0.218 & 0.427 & 0.591 & 0.289 & 0.459 \\
				&\multicolumn{1}{c|}{}& 720   & \textbf{0.071} & \textbf{0.203} & {0.073} & {0.206} & 0.080 & 0.210 & 0.102 & 0.250 & 0.110 & 0.267 & 0.438 & 0.586 & 0.430 & 0.579 \\
				\cline{2-17}
				&\multirow{4}*{\rotatebox{90}{ETTm2}}& 96    & \textbf{0.063} & \textbf{0.181} & 0.065 & 0.186 & \textbf{0.063} & {0.183} & 0.067 & 0.198 & 0.065 & 0.189 & 0.088 & 0.225 & 0.075 & 0.208 \\
				&\multicolumn{1}{c|}{}& 192   & \textbf{0.090} & \textbf{0.224} & 0.094 & 0.231 & {0.092} & {0.227} & 0.102 & 0.245 & 0.118 & 0.256 & 0.132 & 0.283 & 0.129 & 0.275 \\
				&\multicolumn{1}{c|}{}& 336   & \textbf{0.116} & \textbf{0.256} & 0.120 & 0.265 & {0.119} & {0.261} & 0.130 & 0.279 & 0.154 & 0.305 & 0.180 & 0.336 & 0.154 & 0.302 \\
				&\multicolumn{1}{c|}{}& 720   & \textbf{0.170} & \textbf{0.318} & {0.171} & 0.322 & 0.175 & {0.320} & 0.178 & 0.325 & 0.182 & 0.335 & 0.300 & 0.435 & 0.160 & 0.321 \\
				\cline{2-17}
				&\multicolumn{2}{c|}{Count}&13&11&9&4&2&1&0&0&0&0&0&0&0&0\\
				\cline{2-17}
			\end{tabular}
		}
		\vskip -0.15in
	\end{table*}
	
	\textbf{Univarite Time-series Forecasting: Evaluation Summary}
	
	Table \ref{tab::univariate} summarize the univariate evaluation results of all the methods on ETT datasets, which is the univariate series that we are trying to forecast. We cite the baseline results from PatchTST \cite{PatchTST} and DLinear\cite{dlinear}.Among the 16 tests on the ETT dataset, overall, FPN-fusion achieved 13 optima in terms of mse and 11 optima in terms of mae. 
	\begin{itemize}
		\item Compared with DLiner, FPN-fusion achieved a maximum reduction of 54\% in mse, an average decrease of 11.0\%, a maximum reduction of 35.0\% in mae, and an average decrease of 5.5\%.
		\item  Compared with PatchTST, FPN-fusion has 7 leading items and 6 identical items in the mse indicator, with only 3 items worse than PatchTST. In the mae indicator, FPN-fusion has 10 leading items, 1 identical item, and 4 items worse than PatchTST
	\end{itemize}
	
	\begin{table*}[htbp]
		\caption{Multivariate long-term forecasting results. We use prediction lengths $T\in \{24, 36, 48, 60\}$ for ILI dataset and $T\in \{96, 192, 336, 720\}$ for the others. The best results are in \textbf{bold} and the second best are \uline{underlined}.}
		\label{tab:multivarite}
		\vskip 0.15in
		\centering
		\resizebox{\linewidth}{!}{
			\begin{tabular}{cc|c|cc|cc|cc|cc|cc|cc|cc}
				\cline{2-17}
				&\multicolumn{2}{c|}{Models}& \multicolumn{2}{c|}{FPN-fusion}& \multicolumn{2}{c|}{PatchTST}&  \multicolumn{2}{c|}{DLinear}& \multicolumn{2}{c|}{FEDformer}& \multicolumn{2}{c|}{Autoformer}& \multicolumn{2}{c|}{Informer}& \multicolumn{2}{c}{Pyraformer} \\
				\cline{2-17}
				&\multicolumn{2}{c|}{Metric}&MSE&MAE&MSE&MAE&MSE&MAE&MSE&MAE&MSE&MAE&MSE&MAE&MSE&MAE\\
				\cline{2-17}
				&\multirow{4}*{\rotatebox{90}{Weather}}& 96    & \textbf{0.143} & \uline{0.208}& \uline{0.147} & \textbf{0.198} & 0.176 & 0.237 & 0.238 & 0.314 & 0.249 & 0.329 & 0.354 & 0.405 & 0.896 & 0.556  \\
				&\multicolumn{1}{c|}{}& 192   & \textbf{0.187} & \textbf{0.234} &  \uline{0.190} & \uline{0.240} & 0.220 & 0.282 & 0.275 & 0.329 & 0.325 & 0.370 & 0.419 & 0.434 & 0.622 & 0.624 \\
				&\multicolumn{1}{c|}{}& 336   & \textbf{0.238} & \textbf{0.274} & \uline{0.242} & \uline{0.282} & 0.265 & 0.319 & 0.339 & 0.377 & 0.351 & 0.391 & 0.583 & 0.543  & 0.739 & 0.753 \\
				&\multicolumn{1}{c|}{}& 720   & \textbf{0.304}& \textbf{0.325} & \textbf{0.304} & \uline{0.328} & 0.323 & 0.362 & 0.389 & 0.409 & 0.415 & 0.426 & 0.916 & 0.705 & 1.004 & 0.934 \\
				\cline{2-17}
				&\multirow{4}*{\rotatebox{90}{Traffic}}& 96    & \uline{0.408} & \uline{0.273} &  \textbf{0.360} & \textbf{0.249} & 0.410 & 0.282 & 0.576 & 0.359 & 0.597 & 0.371 & 0.733 & 0.410 & 2.085 & 0.468  \\
				&\multicolumn{1}{c|}{} & 192  & \uline{0.419} & \uline{0.286}  & \textbf{0.379} & \textbf{0.256} & 0.423 & 0.287 & 0.610 & 0.380 & 0.607 & 0.382 & 0.777 & 0.435 & 0.867 & 0.467 \\
				&\multicolumn{1}{c|}{}& 336  & \uline{0.422} & \uline{0.285} &  \textbf{0.392} & \textbf{0.264} & 0.436 & 0.296 & 0.608 & 0.375 & 0.623 & 0.387 & 0.776 & 0.434 & 0.869 & 0.469  \\
				&\multicolumn{1}{c|}{}& 720  & \uline{0.443} & \uline{0.299} &  \textbf{0.432} & \textbf{0.286} & 0.466 & 0.315 & 0.621 & 0.375 & 0.639 & 0.395 & 0.827 & 0.466 & 0.881 & 0.473  \\
				\cline{2-17}
				&\multirow{4}*{\rotatebox{90}{Electricity}}& 96   & \uline{0.132} & \uline{0.229} &  \textbf{0.129} & \textbf{0.222} & 0.140 & 0.237 & 0.186 & 0.302 & 0.196 & 0.313 & 0.304 & 0.393 & 0.386 & 0.449  \\
				&\multicolumn{1}{c|}{}& 192  & \uline{0.146} & \uline{0.244} & \textbf{0.141} & \textbf{0.241} & 0.153 & 0.249 & 0.197 & 0.311 & 0.211 & 0.324 & 0.327 & 0.417 & 0.386 & 0.443  \\
				&\multicolumn{1}{c|}{}& 336  & \textbf{0.162} & \uline{0.262}  & \uline{0.163} & \textbf{0.259} & 0.169 & 0.267 & 0.213 & 0.328 & 0.214 & 0.327 & 0.333 & 0.422 & 0.378 & 0.443  \\
				&\multicolumn{1}{c|}{}& 720  & \uline{0.200} & \uline{0.297} &  \textbf{0.197} & \textbf{0.290} & 0.203 & 0.301 & 0.233 & 0.344 & 0.236 & 0.342 & 0.351 & 0.427 & 0.376 & 0.445  \\
				\cline{2-17}
				&\multirow{4}*{\rotatebox{90}{ILI}}& 24   & \uline{1.696} & \uline{0.789} &  \textbf{1.281} & \textbf{0.704} & 2.215 & 1.081 & 2.624 & 1.095 & 2.906 & 1.182 & 4.657 & 1.449  & 1.420 & 2.012  \\
				&\multicolumn{1}{c|}{} & 36    & \uline{1.693} & \uline{0.811}  & \textbf{1.251}& \textbf{0.752} & 1.963 & 0.963 & 2.516 & 1.021 & 2.585 & 1.038 & 4.650 & 1.463 & 7.394 & 2.031\\
				&\multicolumn{1}{c|}{}& 48   & \uline{1.867} & \uline{0.881} & \textbf{1.673} &\textbf{0.854} & 2.130 & 1.024 & 2.505 & 1.041 & 3.024 & 1.145 & 5.004 & 1.542 & 7.551 & 2.057  \\
				&\multicolumn{1}{c|}{}& 60   & \textbf{1.421} & \textbf{0.747}  & \uline{1.526} & \uline{0.795} & 2.368 & 1.096 & 2.742 & 1.122 & 2.761 & 1.114 & 5.071 & 1.543 & 7.662 & 2.100 \\
				\cline{2-17}
				&\multirow{4}*{\rotatebox{90}{ETTh1}}& 96   & \textbf{0.368} & \textbf{0.394} & \uline{0.370} & {0.400} & {0.375} & \uline{0.399} & 0.376 & 0.415 & 0.435 & 0.446 & 0.941 & 0.769 & 0.664 & 0.612  \\
				&\multicolumn{1}{c|}{}& 192  & \uline{0.406} & \uline{0.417} & {0.413} & {0.431} & \textbf{0.405} & \textbf{0.416} & 0.423 & 0.446 & 0.456 & 0.457 & 1.007 & 0.786 & 0.790 & 0.681  \\
				&\multicolumn{1}{c|}{}& 336  & \textbf{0.408} & \textbf{0.425} & \uline{0.422} & \uline{0.440} & 0.439 & 0.443 & 0.444 & 0.462 & 0.486 & 0.487 & 1.038 & 0.784 & 0.891 & 0.738  \\
				&\multicolumn{1}{c|}{}& 720  & \uline{0.458} & \textbf{0.462}  & \textbf{0.447} &  \textbf{0.468} & 0.472 & 0.490 & 0.469 & 0.492 & 0.515 & 0.517 & 1.144 & 0.857 & 0.963 & 0.782 \\
				\cline{2-17}
				&\multirow{4}*{\rotatebox{90}{ETTh2}}& 96  & \uline{0.279} & \textbf{0.333}   & \textbf{0.274} & \uline{0.336} & 0.289 & 0.353 & 0.332 & 0.374 & 0.332 & 0.368 & 1.549 & 0.952 & 0.645 & 0.597\\
				&\multicolumn{1}{c|}{}& 192  & \uline{0.343} & \uline{0.395} &  \textbf{0.339} & \textbf{0.379} & 0.383 & 0.418 & 0.407 & 0.446 & 0.426 & 0.434 & 3.792 & 1.542 & 0.788 & 0.683\\
				&\multicolumn{1}{c|}{}& 336  & \uline{0.379} & \uline{0.423}  & \textbf{0.331} & \textbf{0.380} & 0.448 & 0.465 & 0.400 & 0.447 & 0.477 & 0.479 & 4.215 & 1.642 & 0.907 & 0.747\\
				&\multicolumn{1}{c|}{}& 720   & \uline{0.446} & \uline{0.464}   & \textbf{0.379} & \textbf{0.422} & 0.605 & 0.551 & 0.412 & 0.469 & 0.453 & 0.490 & 3.656 & 1.619 & 0.963 & 0.783\\
				\cline{2-17}
				&\multirow{4}*{\rotatebox{90}{ETTm1}}& 96   & \textbf{0.286} & \textbf{0.335} &\uline{0.290} & \uline{0.342} & 0.299 & {0.343} & 0.326 & 0.390 & 0.510 & 0.492 & 0.626 & 0.560 & 0.543 & 0.510  \\
				&\multicolumn{1}{c|}{}& 192  & \uline{0.330} & \textbf{0.359}  & \textbf{0.328} & 0.365 & \uline{0.335} &{0.365} & 0.365 & 0.415 & 0.514 & 0.495 & 0.725 & 0.619 & 0.557 & 0.537 \\
				&\multicolumn{1}{c|}{}& 336  & \uline{0.368} & \textbf{0.380} & \textbf{0.361} & {0.393} & \uline{0.369} & {0.386} & 0.392 & 0.425 & {0.510} & {0.492} & 1.005 & 0.741 & 0.754 & 0.655  \\
				&\multicolumn{1}{c|}{}& 720  & \uline{0.425} & \textbf{0.413} & \uline{0.416} & 0.419 & 0.425 & {0.421} & 0.446 & 0.458 & 0.527 & 0.493 & 1.133 & 0.845 & 0.908 & 0.724  \\
				\cline{2-17}
				&\multirow{4}*{\rotatebox{90}{ETTm2}} & 96  & \uline{0.163} & \textbf{0.250}  & \textbf{0.162} & \uline{0.254} & 0.167 & 0.260 & 0.180 & 0.271 & 0.205 & 0.293 & 0.355 & 0.462& 0.435 & 0.507  \\
				&\multicolumn{1}{c|}{}& 192  & \textbf{0.216} & \textbf{0.287} & \uline{0.216} & \uline{0.293} & 0.224 & 0.303 & 0.252 & 0.318 & 0.278 & 0.336 & 0.595 & 0.586 & 0.730 & 0.673  \\
				&\multicolumn{1}{c|}{}& 336  & \uline{0.271} & \textbf{0.324}  & \textbf{0.269} & \uline{0.329} & 0.281 & 0.342 & 0.324 & 0.364 & 0.343 & 0.379 & 1.270 & 0.871 & 1.201 & 0.845  \\
				&\multicolumn{1}{c|}{}& 720 & \uline{0.360} & \uline{0.389}  & \textbf{0.350} & \textbf{0.380} & 0.397 & 0.421 & 0.410 & 0.420 & 0.414 & 0.419 & 3.001 & 1.267 & 3.625 & 1.451  \\
				\cline{2-17}
				&\multicolumn{2}{c|}{Count}&10&15&20&17&1&1&0&0&0&0&0&0&0&0\\
				\cline{2-17}
			\end{tabular}
		}
		\vskip -0.15in
	\end{table*}
	
	\textbf{Multivariate Time-series Forecasting: Evaluation Summary}
	
	Table \ref{tab:multivarite} presents the comprehensive evaluation results of various methods on eight datasets. The FPN-fusion model demonstrated outstanding performance, with a total of 10 best mean squared error (MSE) scores and 15 best mean absolute error (MAE) results across the 32 tests. Compared to DLinear, FPN-fusion exhibited an average improvement of 16.8\% in MSE and 11.8\% in MAE.
	
	\begin{itemize}
		\item On the weather dataset, which exhibits strong periodicity, FPN-fusion and patchTST achieved superior performance, with a 1.0\% decrease in MSE and a 0.4\% decrease in MAE compared to DLinear. Notably, these methods outperformed DLinear by 12.6\% and 15.1\% in terms of MAE.
		\item On traffic and electricity datasets, FPN-fusion demonstrated a consistent improvement of approximately 2\% in MSE and 3.3\% in MAE compared to DLinear. The ILI dataset showcased FPN-fusion's prowess, with the best results for a prediction horizon of 60. Compared to PatchTST, FPN-fusion achieved a 7.4\% decrease in MSE and a 6.4\% decrease in MAE. Against DLinear, the improvements were even more substantial, with a 66.6\% decrease in MSE and a 46.7\% decrease in MAE.
		\item In the ETT dataset, FPN-fusion exhibited a strong performance, with 11 optimal MAE results and 4 optimal MSE outcomes. Notably, on the test instance with a prediction length of 192 on ETTh1, FPN-fusion exhibited a 3.4\% decrease in MSE and a 3.5\% decrease in MAE compared to PatchTST.
	\end{itemize}
	
	\subsection{Ablation Study}
	To evaluate the FPN module and fusion mechanism in our FPN fusion algorithm, we conduct an ablation study by comparing Linear, FPNLinear, FPNMLinear, FPN fusion, NLinear, and DLiner models. The performance is averaged over five folds for all evaluation metrics.
	\begin{table*}[htbp]
		\caption{Ablation Study:multivariate long-term forecasting results. The best results are in \textbf{bold} }
		\label{tab:Ablation Study}
		\vskip 0.15in
		\centering
		\resizebox{\linewidth}{!}{
			\begin{tabular}{c c|c|c c|c c|c c|c c|c c}
				\cline{2-13}
				&\multicolumn{2}{c|}{Models}& \multicolumn{2}{c|}{FPN-fusion}& \multicolumn{2}{c|}{FPNMLinear}&  \multicolumn{2}{c|}{FPNLinear}& \multicolumn{2}{c|}{DLinear}& \multicolumn{2}{c}{NLinear} \\
				\cline{2-13}
				&\multicolumn{2}{c|}{Metric}&MSE&MAE&MSE&MAE&MSE&MAE&MSE&MAE&MSE&MAE\\
				\cline{2-13}
				&\multirow{4}*{\rotatebox{90}{Weather}}& 96    & \textbf{0.143} & \textbf{0.208} &\textbf{0.143} & \textbf{0.208} & 0.145 & 0.209 & 0.176 & 0.237 & 0.182 &0.232\\
				&\multicolumn{1}{c|}{}& 192   & \textbf{0.187} & \textbf{0.234} & \textbf{0.187} & {0.252}  & 0.190 & 0.258 & 0.220 & 0.282  & 0.225 &0.269 \\
				&\multicolumn{1}{c|}{}& 336   & \textbf{0.238} & \textbf{0.274} & 0.240 & 0.297 & 0.242 & 0.299 & 0.265 & 0.319 &0.271 &0.301\\
				&\multicolumn{1}{c|}{}& 720   & \textbf{0.304}& \textbf{0.325} & 0.316 & 0.359 & 0.316 & 0.358  & 0.323 & 0.362 &0.338 &0.348\\
				\cline{2-13}
				&\multirow{4}*{\rotatebox{90}{Traffic}}& 96    & \textbf{0.408} & \textbf{0.273}   & 0.411 & 0.282 & 0.409 & 0.282 & 0.410 & 0.282&  0.410 &0.279\\
				&\multicolumn{1}{c|}{} & 192  & \textbf{0.419} & {0.286}   & 0.420 & 0.294 & 0.421 & 0.295  & 0.423 & 0.287& 0.423 &\textbf{0.284} \\
				&\multicolumn{1}{c|}{}& 336  & \textbf{0.422} & \textbf{0.285}   & 0.436 & 0.295 & 0.436 & 0.292 & 0.436 & 0.296& 0.435 &0.290 \\
				&\multicolumn{1}{c|}{}& 720  & \textbf{0.443} & \textbf{0.299} & 0.458 & 0.303 & 0.460 & 0.312  & 0.466 & 0.315 & 0.464 &0.307\\
				\cline{2-13}
				&\multirow{4}*{\rotatebox{90}{Electricity}}& 96   & \textbf{0.132} & \textbf{0.229} & 0.133 & 0.230 & 0.133 & 0.230 & 0.140 & 0.237 &  0.141& 0.237  \\
				&\multicolumn{1}{c|}{}& 192  & \textbf{0.146} & \textbf{0.244}& 0.148 & \textbf{0.244} & 0.148 & 0.245  & 0.153 & 0.249 & 0.154 &0.248 \\
				&\multicolumn{1}{c|}{}& 336  & \textbf{0.162} & \textbf{0.262}& 0.163 & \textbf{0.262} & 0.163 & 0.263  & 0.169 & 0.267  &0.171 &0.265\\
				&\multicolumn{1}{c|}{}& 720  & \textbf{0.200} & \textbf{0.297}  & \textbf{0.200} & \textbf{0.297} & \textbf{0.200} & \textbf{0.297}   & 0.203 & 0.301&  0.210 &\textbf{0.297}\\
				\cline{2-13}
				&\multirow{4}*{\rotatebox{90}{ILI}}& 24   & {1.696} & \textbf{0.789}& 2.097 & 0.968 & 2.115 & 0.951  & 2.215 & 1.081 &\textbf{1.683} &0.858\\
				&\multicolumn{1}{c|}{} & 36    & \textbf{1.693} & \textbf{0.811} & 2.108 & 0.977 & 2.141 & 0.982  & 1.963 & 0.963&1.703 &0.859\\
				&\multicolumn{1}{c|}{}& 48   & {1.867} & \textbf{0.881} & 2.246 & 1.025 & 2.247 & 1.036  & 2.130 & 1.024  & \textbf{1.719} &0.884\\
				&\multicolumn{1}{c|}{}& 60   & \textbf{1.421} & \textbf{0.747}  & 2.198 & 1.023 & 2.170 & 1.013 & 2.368 & 1.096&1.819 &0.917 \\
				\cline{2-13}
				&\multirow{4}*{\rotatebox{90}{ETTh1}}& 96   & \textbf{0.368} & {0.394}  & 0.373 & 0.398 & 0.380 &\textbf{0.392}& {0.375} & {0.399}&0.374 &0.394 \\
				&\multicolumn{1}{c|}{}& 192  & {0.406} & {0.417}  & 0.406 & \textbf{0.415} & 0.406  & {0.421}&\textbf{0.405} &0.416& 0.408& \textbf{0.415} \\
				&\multicolumn{1}{c|}{}& 336  & \textbf{0.408} & \textbf{0.425} & 0.427 & 0.435 & 0.454 & 0.445 & 0.439 & 0.443  & 0.429 &0.427 \\
				&\multicolumn{1}{c|}{}& 720  & {0.458} & {0.462}    & 0.471 & 0.497 & 0.473 & 0.483 & 0.472 & 0.490&\textbf{0.440} &\textbf{0.453}\\
				\cline{2-13}
				&\multirow{4}*{\rotatebox{90}{ETTh2}}& 96  & {0.279} & \textbf{0.333}  & 0.290 & 0.341 & 0.289 & 0.345  & 0.289 & 0.353&\textbf{0.277} &0.338\\
				&\multicolumn{1}{c|}{}& 192  & \textbf{0.343} & {0.395}  & 0.385 &0.416& 0.382 & 0.419 & 0.383 &0.418& 0.344 &\textbf{0.381}\\
				&\multicolumn{1}{c|}{}& 336  & {0.379} & {0.423}  &  0.453 &0.478 &  0.461 &0.483  & 0.448 &0.465& \textbf{0.357} &\textbf{0.400} \\
				&\multicolumn{1}{c|}{}& 720   & {0.446} & {0.464}   & 0.412 & 0.469 & 0.453 & 0.490  & 0.605 & 0.551&\textbf{0.394} &\textbf{0.436}\\
				\cline{2-13}
				&\multirow{4}*{\rotatebox{90}{ETTm1}}& 96   & \textbf{0.286} & \textbf{0.335} & 0.287 & 0.335 & 0.287 & 0.335  & 0.299 & {0.343} &0.306 &0.348 \\
				&\multicolumn{1}{c|}{}& 192  & \textbf{0.327} & \textbf{0.359} & 0.328 & \textbf{0.359} & 0.328 & \textbf{0.359} & {0.335} &{0.365} & 0.349 &0.375  \\
				&\multicolumn{1}{c|}{}& 336  & \textbf{0.368} & \textbf{0.380}& 0.369 & 0.384 & {0.383} & {0.398} & {0.369} & {0.386} & 0.375 &0.388 \\
				&\multicolumn{1}{c|}{}& 720  & \textbf{0.425} & \textbf{0.415}  & \textbf{0.425}  & \textbf{0.415}  & \textbf{0.425}  & 0.416 &\textbf{0.425}  & {0.421}&  0.433 &0.422\\
				\cline{2-13}
				&\multirow{4}*{\rotatebox{90}{ETTm2}} & 96  & \textbf{0.163} & \textbf{0.250} & 0.168 & 0.262 & 0.173 & 0.264  & 0.167 & 0.260 &0.167 &0.255\\
				&\multicolumn{1}{c|}{}& 192  & \textbf{0.216} & \textbf{0.287}  & 0.236 & 0.326 & 0.225 & 0.308  & 0.224 & 0.303& 0.221 &0.293\\
				&\multicolumn{1}{c|}{}& 336  & \textbf{0.271} & \textbf{0.324}  & 0.283 & 0.364 & 0.303 & 0.357  & 0.281 & 0.342& 0.274 &0.327  \\
				&\multicolumn{1}{c|}{}& 720 & \textbf{0.361} & {0.386}   & 0.392 & 0.410 & 0.363 & 0.399 & 0.397 & 0.421& 0.368 &\textbf{0.384}\\
				\cline{2-13}
				&\multicolumn{2}{c|}{Avg}&0.480	&0.390&	0.550&	0.426&	0.554&	0.426&	0.562&	0.437&	0.496&	0.403\\
				\cline{2-13}
				&\multicolumn{2}{c|}{Count}&25&24&4&7&2&3&2&0&6&8\\
				\cline{2-13}
				&\multicolumn{2}{c|}{Improvement}&\multicolumn{2}{c|}{---}&14.6\%&9.0\%&15.3\%&	9.2\%&	16.8\%&	11.8\%&	3.1\%&	3.4\%\\
				\cline{2-13}
			\end{tabular}
		}
		\vskip -0.15in
	\end{table*}
	
	Table \ref*{tab:Ablation Study} presents the comparative results of these models, with FPN fusion achieving optimal performance in 32 tests, attaining 25 minimum mean squared error (MSE) and 24 minimum mean absolute error (MAE) instances. Compared to the DLinear model, FPN fusion exhibits a noteworthy improvement of 16.8\% in MSE and 11.8\% in MAE. The FPN method outperforms FPNLiner in feature extraction, and FPNMLinear further enhances this by offering a modest improvement over FPNLiner.
	
	The comparison between FPNLinear and FPNMLinear against DLinear reveals that FPN's feature extraction approach consistently surpasses the decomposition-based DLinear model, particularly in handling complex data with trend and fluctuation components.
	
	Moreover, the FPN fusion module exhibits a substantial enhancement over FPNMLinear, highlighting its effectiveness in mitigating duplicate trend features and enhancing the representation of fine-grained information.
	
	\subsection{model efficiency}
	we analyze the efficiency of FPN fusion by examining computational complexity, model parameters, and GPU memory usage, comparing it with reference models such as DLiner, PatchTST, Autoformer, and Informer. This evaluation provides insights into the trade-offs between model performance and computational efficiency.
	
	\begin{table}[htbp]
		\caption{Compare the parameter quantity and resource consumption of the model under the look-back window $ L = 336 $ , prediction length  $T = 96 $ and $batch\_size = 32$ on the ETTh2. MACS are the number of multiply-accumulate operations.}
		\label{tab:parameter}
		\vskip 0.15in
		\begin{center}
			\begin{small}
				\begin{tabular}{l|ccc}
					\toprule
					Method & MACs & Parameter  & Memory \\
					\midrule
					FPN-fusion   & 13.56M &  0.42M  & 20.39M \\
					\hline
					DLinear & 14.52M & 0.45M  & 20.55M \\
					\hline
					PatchTST & 164.97M & 0.46M  & 20.72M  \\
					\hline
					Autoformer & 90517.61M & 10.54M  & 119.39M   \\
					\hline
					Informer & 79438.08M & 11.33M  & 126.22M  \\
					\bottomrule
				\end{tabular}
			\end{small}
		\end{center}
		\vskip -0.1in
	\end{table}
	
	Table \ref{tab:parameter} presents a detailed comparison of model efficiency, focusing on parameters, computational complexity (measured by Multiply-Accumulate Operations, or MACS), and GPU memory usage. The experiments were conducted with a fixed look-back window of 336, a prediction length of 96, and a batch size of 32 on the ETTh2 dataset. The results are averaged over five runs to ensure fairness in the comparison.
	
	\begin{itemize}
		\item FPN-fusion and DLinear exhibit comparable computational efficiency, with FPN-fusion demonstrating a slight advantage. PatchTST, which relies on patch-based methods, has a significantly higher computational cost, approximately 12.1 times that of FPN-fusion. By employing a smaller pooling kernel, FPN-fusion reduces computational requirements by 6.6\%, further enhancing its efficiency.
		
		\item In terms of model size and memory footprint, FPN-fusion is on par with DLinear and PatchTST. Given its superior performance and resource efficiency, FPN-fusion is a more practical choice for real-world applications, especially in univariate and multivariate time series forecasting tasks where resource constraints are prevalent.
	\end{itemize}
	\subsection{Results and Discussion}
	The results indicate that FPN fusion outperforms competitive methods in prediction accuracy while maintaining linear computational complexity. The improvement in performance is attributed to FPN feature extraction and effective multi-level fusion, which allows for better handling of complex patterns in the data. In addition, linear complexity ensures that the model is scalable to large datasets, making it a practical choice for real-world applications.
	
	\section{Conclusion and Future Work}
	In summary, FPN fusion provides a promising solution for effective and accurate time series prediction challenges. We have achieved a model that balances computational efficiency and predictive ability through FPN feature decomposition structure and multi-layer fusion structure. Future work will explore further optimization techniques and extensions to other fields, such as multi-step prediction and online learning.
	\bibliographystyle{unsrt}
	\bibliography{FPN-fusion}
\newpage
\section*{NeurIPS Paper Checklist}

\begin{enumerate}

\item {\bf Claims}
    \item[] Question: Do the main claims made in the abstract and introduction accurately reflect the paper's contributions and scope?
    \item[] Answer: \answerYes{} 
    \item[] Justification: The paper includes our mathematical formulation, quantitative experimental results, and qualitative visual examples that reflect and justify the claims in our abstract and introduction.
    \item[] Guidelines:
    \begin{itemize}
        \item The answer NA means that the abstract and introduction do not include the claims made in the paper.
        \item The abstract and/or introduction should clearly state the claims made, including the contributions made in the paper and important assumptions and limitations. A No or NA answer to this question will not be perceived well by the reviewers. 
        \item The claims made should match theoretical and experimental results, and reflect how much the results can be expected to generalise to other settings. 
        \item It is fine to include aspirational goals as motivation as long as it is clear that these goals are not attained by the paper. 
    \end{itemize}

\item {\bf Limitations}
    \item[] Question: Does the paper discuss the limitations of the work performed by the authors?
    \item[] Answer: \answerYes{} 
    \item[] Justification: The discussion section contains a discussion of our method's limitations.
    \item[] Guidelines:
    \begin{itemize}
        \item The answer NA means that the paper has no limitation while the answer No means that the paper has limitations, but those are not discussed in the paper. 
        \item The authors are encouraged to create a separate "Limitations" section in their paper.
        \item The paper should point out any strong assumptions and how robust the results are to violations of these assumptions (e.g., independence assumptions, noiseless settings, model well-specification, asymptotic approximations only holding locally). The authors should reflect on how these assumptions might be violated in practice and what the implications would be.
        \item The authors should reflect on the scope of the claims made, e.g., if the approach was only tested on a few datasets or with a few runs. In general, empirical results often depend on implicit assumptions, which should be articulated.
        \item The authors should reflect on the factors that influence the performance of the approach. For example, a facial recognition algorithm may perform poorly when image resolution is low or images are taken in low lighting. Or a speech-to-text system might not be used reliably to provide closed captions for online lectures because it fails to handle technical jargon.
        \item The authors should discuss the computational efficiency of the proposed algorithms and how they scale with dataset size.
        \item If applicable, the authors should discuss possible limitations of their approach to address problems of privacy and fairness.
        \item While the authors might fear that complete honesty about limitations might be used by reviewers as grounds for rejection, a worse outcome might be that reviewers discover limitations that aren't acknowledged in the paper. The authors should use their best judgment and recognize that individual actions in favor of transparency play an important role in developing norms that preserve the integrity of the community. Reviewers will be specifically instructed to not penalise honesty concerning limitations.
    \end{itemize}

\item {\bf Theory Assumptions and Proofs}
    \item[] Question: For each theoretical result, does the paper provide the full set of assumptions and a complete (and correct) proof?
    \item[] Answer: \answerYes{} 
    \item[] Justification: We state at the beginning of our derivation the only assumption of our method (that conditions are independent), followed by the full derivation of our formulation, with extra intermediate steps in the Appendix.

    \item {\bf Experimental Result Reproducibility}
    \item[] Question: Does the paper fully disclose all the information needed to reproduce the main experimental results of the paper to the extent that it affects the main claims and/or conclusions of the paper (regardless of whether the code and data are provided or not)?
    \item[] Answer: \answerYes{} 
    \item[] Justification: We show how our formulation result can be applied to any discrete generative prior (in addition to specific adaptations to the particular non-autoregressive prior we used). We provide anonymized code for our quantitative experiments alongside clear instructions (README.md) for training and evaluation.

\item {\bf Open access to data and code}
    \item[] Question: Does the paper provide open access to the data and code, with sufficient instructions to faithfully reproduce the main experimental results, as described in supplemental material?
    \item[] Answer: \answerYes{} 
    \item[] Justification: We provide anonymized code for our quantitative experiments alongside clear instructions (README.md) for training and evaluation.

\item {\bf Experimental Setting/Details}
    \item[] Question: Does the paper specify all the training and test details (e.g., data splits, hyperparameters, how they were chosen, type of optimizer, etc.) necessary to understand the results?
    \item[] Answer: \answerYes{} 
    \item[] Justification: Full training and testing details are in appendix. Full implementations of generative models and classifiers are included in code.

\item {\bf Experiment Statistical Significance}
    \item[] Question: Does the paper report error bars suitably and correctly defined or other appropriate information about the statistical significance of the experiments?
    \item[] Answer: \answerYes{} 
    \item[] Justification: For results on generation accuracy (Acc.), we include error bars at $\pm2\sigma$ derived from the number of trials (5000). For FID results we explain why this is not possible

\item {\bf Experiments Compute Resources}
    \item[] Question: For each experiment, does the paper provide sufficient information on the computer resources (type of compute workers, memory, time of execution) needed to reproduce the experiments?
    \item[] Answer: \answerYes{} 
    \item[] Justification: We include the name of the GPU we used in addition to the time of execution for individual experiments and all experiments in total (in days).
    
\item {\bf Code Of Ethics}
    \item[] Question: Does the research conducted in the paper conform, in every respect, with the NeurIPS Code of Ethics \url{https://neurips.cc/public/EthicsGuidelines}?
    \item[] Answer: \answerYes{} 
    \item[] Justification: We do not work with human participants, and all relevant datasets (FFHQ) have been checked for privacy compliance prior to experiments and submission. We do not release any model that could be considered high-risk, and we offer a discussion of broader societal impacts (including bias and misuse) in our discussion section.

\item {\bf Broader Impacts}
    \item[] Question: Does the paper discuss both potential positive societal impacts and negative societal impacts of the work performed?
    \item[] Answer: \answerYes{} 
    \item[] Justification: Our discussion section has a subsection dedicated to potential societal impacts, including both positive and negative.
    
\item {\bf Safeguards}
    \item[] Question: Does the paper describe safeguards that have been put in place for responsible release of data or models that have a high risk for misuse (e.g., pretrained language models, image generators, or scraped datasets)?
    \item[] Answer: \answerNA{} 
    \item[] Justification:  Our contribution does not include new datasets or pre-trained models that pose a risk of misuse.

\item {\bf Licenses for existing assets}
    \item[] Question: Are the creators or original owners of assets (e.g., code, data, models), used in the paper, properly credited and are the license and terms of use explicitly mentioned and properly respected?
    \item[] Answer: \answerYes{} 
    \item[] Justification: Code that we derive from earlier work is properly licensed and referenced (see code LICENSE).

\item {\bf New Assets}
    \item[] Question: Are new assets introduced in the paper well documented and is the documentation provided alongside the assets?
    \item[] Answer: \answerYes{} 
    \item[] Justification: We provide anonymized code for our quantitative experiments alongside clear instructions for training and evaluation.

\item {\bf Crowdsourcing and Research with Human Subjects}
    \item[] Question: For crowdsourcing experiments and research with human subjects, does the paper include the full text of instructions given to participants and screenshots, if applicable, as well as details about compensation (if any)? 
    \item[] Answer: \answerNA{} 
    \item[] Justification: No human subjects or crowdsourcing were involved in this research.

\item {\bf Institutional Review Board (IRB) Approvals or Equivalent for Research with Human Subjects}
    \item[] Question: Does the paper describe potential risks incurred by study participants, whether such risks were disclosed to the subjects, and whether Institutional Review Board (IRB) approvals (or an equivalent approval/review based on the requirements of your country or institution) were obtained?
    \item[] Answer: \answerNA{} 
    \item[] Justification: No human subjects were involved in this research.

\end{enumerate}

\end{document}